\documentclass[sigconf]{acmart}
\newcommand{\proposed}{DHAG}
\newcommand*{\defeq}{\stackrel{\text{def}}{=}}
\usepackage{subfig}
\usepackage{amsfonts}
\usepackage{algorithmic}
\usepackage{algorithm}
\usepackage{enumitem}
\usepackage{balance}
\AtBeginDocument{%
  }

\copyrightyear{2024}
\acmYear{2024}
\setcopyright{acmlicensed}\acmConference[CIKM '24]{Proceedings of the 33rd
ACM International Conference on Information and Knowledge
Management}{October 21--25, 2024}{Boise, ID, USA}
\acmBooktitle{Proceedings of the 33rd ACM International Conference on
Information and Knowledge Management (CIKM '24), October 21--25, 2024,
Boise, ID, USA}
\acmDOI{10.1145/3627673.3679623}
\acmISBN{979-8-4007-0436-9/24/10}

\begin{document}

%%
%% The "title" command has an optional parameter,
%% allowing the author to define a "short title" to be used in page headers.
\title{Enhancing Anomaly Detection via Generating \\ Diversified and Hard-to-distinguish Synthetic Anomalies}

%%
%% The "author" command and its associated commands are used to define
%% the authors and their affiliations.
%% Of note is the shared affiliation of the first two authors, and the
%% "authornote" and "authornotemark" commands
%% used to denote shared contribution to the research.

\author{Hyuntae Kim}
\orcid{0009-0005-2247-9571}
\affiliation{%
    % \department{Department of Artificial Intelligence}
    \institution{Chung-Ang University}
    \city{Seoul}
    \country{Republic of Korea}
    }
\email{soodaman97@cau.ac.kr}

\author{Changhee Lee}
\orcid{0000-0002-8681-4739}
\authornote{Corresponding author.}
\affiliation{%
    % \department{Department of Artificial Intelligence}
    \institution{Korea University}
    \city{Seoul}
    \country{Republic of Korea}
    }
\email{chl8856@gmail.com}

%%
%% By default, the full list of authors will be used in the page
%% headers. Often, this list is too long, and will overlap
%% other information printed in the page headers. This command allows
%% the author to define a more concise list
%% of authors' names for this purpose.
\renewcommand{\shortauthors}{Hyuntae Kim and Changhee Lee}

%%
%% The abstract is a short summary of the work to be presented in the
%% article.
\begin{abstract}
Unsupervised anomaly detection is a daunting task, as it relies solely on normality patterns from the training data to identify unseen anomalies during testing. Recent approaches have focused on leveraging domain-specific transformations or perturbations to generate synthetic anomalies from normal samples. The objective here is to acquire insights into normality patterns by learning to differentiate between normal samples and these crafted anomalies. However, these approaches often encounter limitations when domain-specific transformations are not well-specified such as in tabular data, or when it becomes trivial to distinguish between them. To address these issues, we introduce a novel domain-agnostic method that employs a set of conditional perturbators and a discriminator. The perturbators are trained to generate input-dependent perturbations, which are subsequently utilized to construct synthetic anomalies, and the discriminator is trained to distinguish normal samples from them. We ensure that the generated anomalies are both diverse and hard to distinguish through two key strategies: i) directing perturbations to be orthogonal to each other and ii) constraining perturbations to remain in proximity to normal samples. Throughout experiments on real-world datasets, we demonstrate the superiority of our method over state-of-the-art benchmarks, which is evident not only in image data but also in tabular data, where domain-specific transformation is not readily accessible. Additionally, we empirically confirm the adaptability of our method to semi-supervised settings, demonstrating its capacity to incorporate supervised signals to enhance anomaly detection performance even further. 
\end{abstract}




%%
%% This command processes the author and affiliation and title
%% information and builds the first part of the formatted document.
\maketitle

\section{Introduction}
Unsupervised anomaly detection involves identifying abnormal samples whose patterns deviate from the expected normal behaviors learned from data composed solely of normal samples. This is a fundamental problem in various domains, including healthcare, manufacturing, and finance. 
In healthcare, for instance, diseases often manifest through a broad spectrum of patient features, leading to heterogeneous patterns that can be distinguished from the healthy population. Recognizing such deviations is crucial for anticipating adverse disease progressions, empowering healthcare practitioners to make informed clinical decisions \cite{vsabic2021healthcare}.
In manufacturing, detecting anomalies in a production process is crucial for achieving high-quality products and minimizing financial losses. This is achieved by precisely identifying deviations from normal patterns captured by industrial sensors, which exhibit complex correlations among multiple variables \cite{maggipinto2022deep}. 
Such numerous applications across various domains have led to a growing demand for anomaly detection methods capable of universally applying to diverse data types spanning different domains while effectively capturing their complex patterns.

Many recent unsupervised anomaly detection methods employ self-supervised learning frameworks to acquire normality patterns, achieved either through the utilization of contrasting learning \cite{DROC,tack2020csi} or auxiliary classification tasks \cite{CutPaste,Draem}.  
The core idea here is to learn representations that preserve relevant information about normal samples or establish a decision boundary capable of identifying normal samples, both by comparing them to their augmented counterparts. This underscores the critical role of augmentation techniques in these methods, as they provide diverse aspects of (potential) abnormal patterns from normal samples. 
Unfortunately, these methods primarily focus on image data by leveraging augmentations specifically designed for images -- such as rotations, reflections, and cropping -- making them unsuitable for general data types across diverse domains. 

Learnable domain-agnostic transformations for unsupervised anomaly detection \cite{PLAD,DROCC} have recently been proposed to overcome the limitation associated with relying on domain-specific (image) transformations. These methods primarily focus on \textit{jointly} acquiring the ability to generate anomalous versions of normal samples while learning the normality patterns by discriminating them from the generated synthetic anomalies.\footnote{It is worth highlighting that none of these methods, including PLAD \cite{PLAD}, DROCC \cite{DROCC}, and ours, can guarantee that the characteristics of real anomalies (unseen during training) will align with those found by synthetically generated anomalies.} 
These methods often encounter two fundamental challenges that complicate anomaly detection during testing: 
First, as the discriminator is designed to detect subtle deviations (i.e., generated by small perturbations) from normal samples, it may become overly sensitive when applied to unseen normal samples in the testing set, leading to unnecessary false alarms. 
Second, due to the concurrent training of the perturbator and discriminator pair, the perturbator may easily collapse to generate perturbations that the discriminator can trivially distinguish. This leads to a degenerate discriminator, rendering the model unsuitable for identifying (unseen) real anomalies during testing.

% \paragraph{\textbf{Contribution.}} % policy about latex 
\paragraph{Contribution.}
To address the above challenges, we introduce a novel perturbation-based unsupervised anomaly detection method that generates synthetic anomalies that are \textit{diversified} in the latent space and \textit{hard-to-distinguish} when used by the discriminator, based on the following two novel strategies:
\begin{itemize}[leftmargin=1.5em]
    \item We align synthetically generated anomalies with normal samples by constructing augmented versions of normal samples. This is achieved by selecting a set of $K$ perturbed samples with the smallest perturbation level as normal, while treating the remaining perturbed samples as abnormal. This further complicates the discriminator's task of distinguishing synthetic anomalies from normal samples by ensuring that synthetic anomalies are closely aligned with the normal data on their latent representations.
    \item We force the multiple perturbators to generate perturbations in orthogonal directions within the latent space. This diversifies the latent representations of the generated anomalies and guides the discriminator to differentiate them from the normal representations in various directions.  While we cannot ensure that the generated perturbations will exhibit similar latent representations to unseen real anomalies, our discriminator learns a more generalizable decision boundary than that with a single perturbator, making it more effective in identifying unseen real-world anomalies.    
\end{itemize}
The effectiveness of the two strategies is well demonstrated through the visualization of the learned representations in Figure \ref{fig:tsne_visualization}. While the generated synthetic anomalies do not perfectly overlap with real anomalies, our strategy makes the perturbations cover a wider range of potential anomaly directions originating from normal samples. Furthermore, by applying $K$ augmentation, the synthetically generated anomalies remain in the vicinity of normal samples, preventing the discriminator from learning trivial distinctions.

Throughout experiments on real-world image and tabular datasets, we demonstrate the remarkable effectiveness of our proposed method by achieving superior anomaly detection performance over state-of-the-art benchmarks. This superiority is evident not only in image data but also in tabular data, where domain-specific transformation is not readily accessible. Moreover, we empirically confirm the adaptability of our method to semi-supervised settings, demonstrating its capacity to incorporate supervised signals from the labeled anomalies to enhance the detection performance even further.

\section{Related Works}
%%%%%%%%%%%%%%%%%%%%%%
% 2. Related Works
%%%%%%%%%%%%%%%%%%%%%%
Unsupervised anomaly detection aims to learn a score function from normal samples capable of identifying (unseen) abnormal samples during testing. 
There are various existing works under different modeling assumptions: i) abnormal samples incur relatively higher reconstruction costs in comparison to normal samples \cite{DAE}, ii) normal data is densely concentrated around the center of a hypersphere \cite{DSVDD}, iii) exposure to samples from (disjoint) auxiliary datasets helps identifying abnormal samples during testing \cite{DOE}, and iv) the distribution of normal data can be modeled using specific types of distributions, e.g., Gaussian mixture models \cite{DAGMM}. 
In this paper, we focus on describing approaches based on \textit{different types of transformations}, that are most closely related to our work.

\paragraph{Domain-Specific Transformations.} 
One line of research has directed its attention toward acquiring representations based on self-supervised learning beneficial for anomaly detection by employing various forms of image transformations \cite{DROC,tack2020csi,CutPaste,Draem}.  
A novel contrastive learning framework has been introduced to extract high-level representations based on distribution-shifting transformations (e.g., rotation and flipping) of a given normal sample, creating its negative pairs \cite{DROC,tack2020csi}. Then, these acquired representations are effectively applied in either learning one-class classifiers \cite{DROC} or establishing a score function for anomaly detection \cite{tack2020csi}. 
Other approaches introduce an auxiliary task of distinguishing normal samples from synthetic anomalies generated based on image-specific augmentations such as cut-out \cite{CutPaste} and random augmentations \cite{Draem}.
However, these methods heavily depend on image-specific augmentations, rendering them inapplicable to general data from other domains.

\paragraph{Domain-Agnostic Transformations.} 
To overcome the limitation due to the reliance on domain-specific (image) transformations, another line of research has focused on acquiring representations for anomaly detection through the utilization of domain-agnostic transformations. 
In \cite{bergman2020goad,qiu2021neural}, the authors introduced multiple transformations to create diverse views of normal samples. More specifically, GOAD \cite{bergman2020goad} randomly initializes a set of affine transformations and utilizes the transformed normal samples in training a classifier based on the triplet loss to distinguish between normal samples and their respective transformed counterparts. NeutraLAD \cite{qiu2021neural} employs a set of learnable neural transformations, jointly trained with an encoder based on the deterministic contrastive loss. This encourages the transformed samples to be similar to the original sample while remaining dissimilar to other transformed versions, thus rendering the learned representations to preserve relevant semantic information of normal samples. 
However, as these approaches depend solely on contrasting the transformed versions of the original samples, they might struggle to acquire meaningful information if the transformation is not adequately initialized or trained. 
Also, these methods cannot be extended to a semi-supervised setting where we are provided with some known anomalies.

Our work is most closely related to  \cite{PLAD,DROCC} which introduce trainable perturbators that generate synthetic anomalies by altering feature values of normal samples. These synthetically generated anomalies are then utilized in training a classifier to obtain a decision boundary capable of distinguishing normal samples from their perturbed counterparts. Specifically, DROCC \cite{DROCC} employs adversarial training to generate anomalous samples based on corresponding normal samples. This process involves iterative updates to the perturbations within a predefined perturbation level using gradient ascent. In contrast, PLAD \cite{PLAD} generates two input-dependent vectors -- i.e., multiplicative and additive perturbations -- which are directly multiplied and added to alter corresponding input (normal) samples, introducing subtle defects like salt and pepper noise.  
While our work also utilizes the classification-based auxiliary task, our proposed perturbations are diverse and hard to distinguish thanks to two key factors: i) simultaneous generation of both normal and abnormal perturbations and ii) utilization of multiple perturbators spanning distinct directions. This helps to learn a better decision boundary by enhancing generalization to unseen normal samples and making the classification task more challenging. Furthermore, the proposed perturbations are applied in the latent space to prevent the generation of trivial defects that can be easily distinguishable within the feature space.

\section{Problem Formulation}
%%%%%%%%%%%%%%%%%%%%%%
% 3. Problem
%%%%%%%%%%%%%%%%%%%%%%
Suppose we have a training set $\mathcal{D}_{tr}=\{x^i\}_{i=1}^{N_{tr}}$ comprising normal samples drawn from an input distribution $\mathcal{P}_X$, i.e., $x^{i}\in \mathcal{X} \sim \mathcal{P}_X$, 
and a testing set $\mathcal{D}_{te}=\{(x^i, y^i)\}_{i=1}^{N_{te}}$ containing abnormal samples that are unlikely drawn from the input distribution i.e., $p_{X}(x^{i}) \approx 0$. 
The label, $y^{i} \in \{0,1\}$, corresponding to $x^i$, denotes whether the sample is normal (i.e., $y^i =0$) or anomalous (i..e, $y^i=1$). 
We consider an unsupervised anomaly detection problem where the goal is to learn a score function, i.e., $S: \mathcal{X} \rightarrow [0,1]$, from the training data that can distinguish unseen abnormal samples in the testing data. It is worth highlighting that we consider scenarios with different data types such as images, i.e., $\mathcal{X} = \mathbb{R}^{W\times H \times C}$, and tabular data, i.e., $\mathcal{X} = \mathbb{R}^{d}$. From this point forward, we will omit the dependency on $i$ when it is clear in the context.

We present our novel unsupervised anomaly detection framework based on \textit{\textbf{D}iversified and \textbf{H}ard-to-distinguish \textbf{A}nomaly \textbf{G}eneration}, which we refer to as \proposed, following recent advances utilizing learnable domain-agnostic perturbations \cite{PLAD}. 
Our goal is to train a discriminator capable of distinguishing synthetically generated abnormal samples, created by perturbing original samples, from unaltered original (normal) samples. 
We encourage the perturbations to have the following two properties: 
First, the perturbations must be subtle to ensure that the generated abnormal samples remain close to the original samples. This complicates the task of distinguishing between the original and perturbed samples, effectively preventing the acquisition of trivial solutions. 
Second, the perturbations must generate anomalous versions of the original data from different perspectives without relying on domain knowledge. This enables us to establish a decision boundary that generalizes well to unseen real anomalies. 

Unfortunately, most previous works fail to address the aforementioned properties as they either focus on random perturbations that are not dependent on the input normal sample \cite{DROCC}, which often results in generating synthetic anomalies that can be easily distinguished by the discriminator, or heavily rely on domain-specific augmentations \cite{DROC,CutPaste}, which are often not accessible in practice.

%% ============================ ALGORITHM 1 ===============================%
\begin{algorithm}[htb!]
\small
        \caption{Pseudo-code for the Training Phase of \proposed}
	\label{alg:DRAG_train}
	\begin{algorithmic}
        \STATE {\bfseries Input:}~~Dataset $\mathcal{D}_{tr}=\{x^i\}_{i=1}^{N_{tr}}$, learning rates $\eta_1, \eta_2 > 0$, \\ mini-batch size $m$, coefficients $\lambda_1, \lambda_2 > 0$, num of perturbators $L$, \\num of augmented version of samples $K$, the permutation function $\pi(\cdot)$
        % \STATE        
        \STATE {\bfseries Initialize:}~~\proposed~parameters $(\theta, \phi, \{ \psi_\ell \}_{\ell=1}^{L})$
        % \STATE
        \REPEAT
        \STATE Sample a mini-batch of $m$ unlabeled samples:~~$\{ x^i \}_{i=1}^m \sim \mathcal{D}_{tr}$
        \FOR{$\ell=1,\cdots, L$} 
            \STATE Generate perturbations:~~$\mathcal{E}=\{ \epsilon_\ell^i \sim g_{\psi_\ell}(x^i) \}_{i=1}^{m}$
            \STATE Permute perturbations in ascending order:~~\\ \qquad \qquad $(\epsilon^{\pi(1)}_{\ell}, \epsilon^{\pi(2)}_{\ell}, \dots, \epsilon^{\pi(m)}_{\ell}) \leftarrow \pi(\mathcal{E})$
            \STATE Separate perturbations in two sets:~~\\ \qquad \qquad $\mathcal{E}^{+}_\ell \defeq \{ \epsilon_{\ell}^{\pi(j)} \}_{j=1}^{K}, \quad  \mathcal{E}^{-}_\ell \defeq \{ \epsilon_{\ell}^{\pi(j)}\}_{j=K+1}^{m}$
            % $\mathcal{E}_+ \defeq \{ \epsilon_{\ell, +}^j \, | \, \| \epsilon_\ell^j \|_2 \leq \delta \}_{j=1}^{K}, \; \mathcal{E}_- \defeq \{ \epsilon_{\ell, -}^j \, | \, \| \epsilon_\ell^j \|_2 > \delta \}_{j=1}^{m-K}\quad( \delta=\| \epsilon_\ell^{\pi(k)} \|_2)$
            \STATE Assign a pseudo label:~~ \qquad 
            \begin{equation} \nonumber %\hskip-150pt
            \tilde{y}^i_\ell =\begin{cases} 
                0~~\text{(normal)} &\text{for}~~\epsilon_{\ell}^{i} \in \mathcal{E}^{+}_\ell \\
                1~~\text{(abnormal)} &\text{for}~~\epsilon_{\ell}^{i} \in \mathcal{E}^{-}_\ell
            \end{cases} 
            ~~~~~\quad \text{for}~~~i=1,\dots,m
            \end{equation}
        \ENDFOR
    \STATE Compute loss functions: \vspace{0mm}
    \begin{align}  \nonumber 
      & \mathcal{L}_{ce} \!= \frac{1}{m} \sum_{i=1}^m  \texttt{CE}(f_\phi (f_\theta(x^i)), 0) +  \frac{1}{L} \sum_{\ell=1}^L \texttt{CE}(f_\phi(f_\theta (x^i) \!+\! \epsilon_{\ell}^i), \tilde{y}_\ell^i)  \\\nonumber 
      & \mathcal{L}_{norm} \!= \frac{1}{m} \sum_{i=1}^m  \frac{1}{L} \sum_{\ell=1}^L  \| \epsilon_{\ell}^i \|_2 ,~~    \mathcal{L}_{div}  \!= \frac{1}{m} \sum_{i=1}^m \frac{1}{L(L\!-\!1)} \!\sum_{\ell \neq k} \texttt{sim}(\epsilon_{\ell}^i, \epsilon_{k}^i) \\ \nonumber 
      & \mathcal{L}_{total} \!= \mathcal{L}_{ce} + \lambda_1 \mathcal{L}_{norm} + \lambda_2 \mathcal{L}_{div} 
   \end{align} 
    \STATE Update the encoder parameters:
        $\theta \leftarrow \theta - \eta_1 \nabla_{\theta}  \mathcal{L}_{total} $
    \STATE Update the perturbator parameters: $\psi_\ell \leftarrow \psi_\ell - \eta_1 \nabla_{\psi_\ell}  \mathcal{L}_{total}$
    \STATE Update the discriminator parameters: $\phi \leftarrow \phi - \eta_2 \nabla_{\phi}  \mathcal{L}_{total}$
    % \STATE
    \UNTIL converge    
	\end{algorithmic}
\end{algorithm}
%% =========================================================================%

%%%%%%%%%%%%%%%%%%%%%%%%%%%%%%%%%%%%%
% Architecture figure
\begin{figure*}[t]
    \centering
    \includegraphics[width=0.80\linewidth,height=180pt, trim=0em 7em 0em 0em, clip]{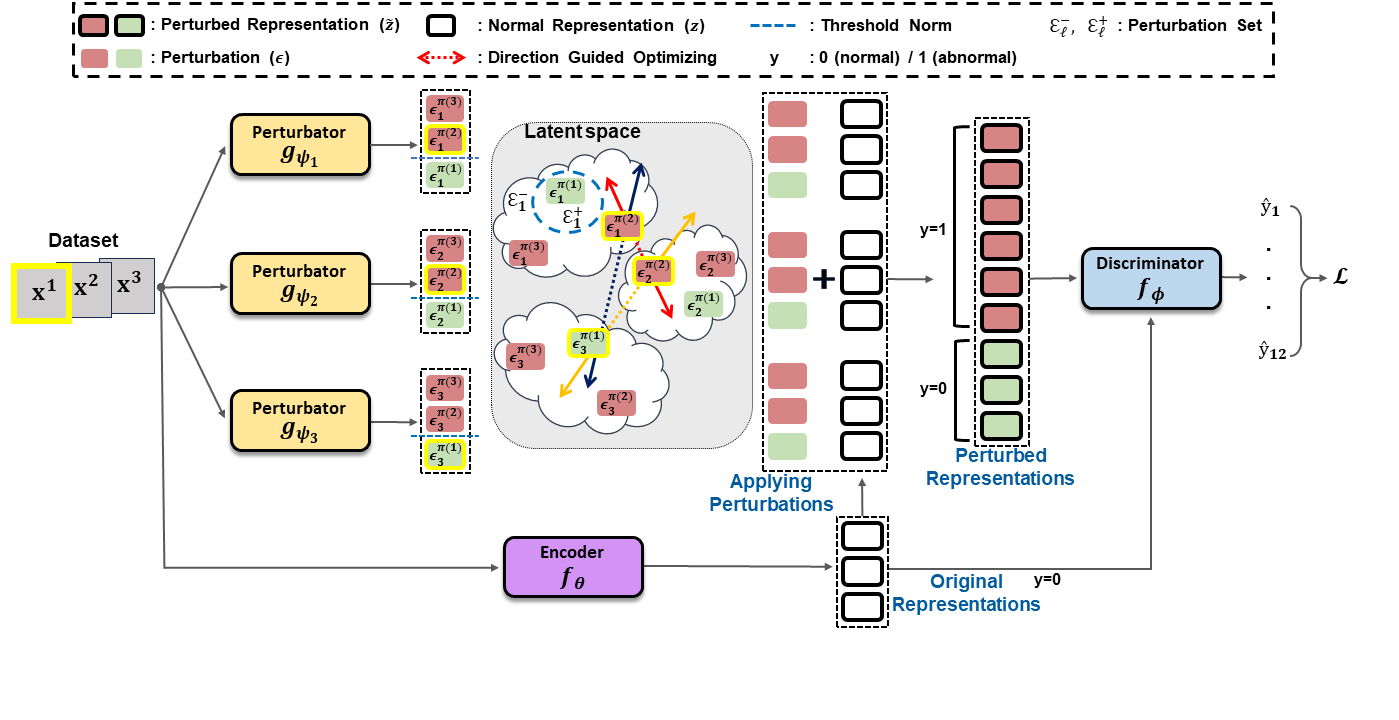}
    \caption{Overall network architecture of our method. We present a toy example with three samples ($m=3$), three perturbators ($L=3$), and the number of augmentation perturbations, $K$, set to 1.}
    \label{fig:architecture}
\end{figure*}
%%%%%%%%%%%%%%%%%%%%%%%%%%%%%%%%%%%%%

\section{Method}
%%%%%%%%%%%%%%%%%%%%%%
% 4. Method
%%%%%%%%%%%%%%%%%%%%%%
For the proposed unsupervised anomaly detection framework, we introduce three networks -- an \textit{encoder}, a set of $L$ \textit{perturbators}, and a \textit{discriminator} -- as shown in Figure \ref{fig:architecture} and Algorithm \ref{alg:DRAG_train}. These components work together to achieve a discriminator that is capable of distinguishing normal samples from synthetically generated abnormal samples that are both diverse and hard to distinguish. We define each component as follows:
\begin{itemize}[leftmargin=1.5em]
    \item The \textit{encoder}, $f_{\theta}:\mathcal{X}\rightarrow \mathcal{Z}$, which transforms complex input features $x \in \mathcal{X}$ into a latent representation $z \in \mathcal{Z}$ where $\mathcal{Z}$ is the latent space, i.e., $z = f_{\theta}(x)$. 
    \item A set of $L$ \textit{perturbators}, $\{g_{\psi_{\ell}} \}_{\ell=1}^{L}$, where each perturbator, $g_{\psi_{\ell}}: \mathcal{X} \rightarrow \mathcal{Z}$, is a stochastic function that generates perturbations in the latent space conditioned on $x$, i.e., $\epsilon_{\ell} \sim  g_{\psi_{\ell}}(x)$. Perturbations are then applied to the latent representation of $x$ to create a latent representation for the synthetically generated anomalous sample, i.e., $\tilde{z}_{\ell} = z + \epsilon_{\ell}$.
    \item The \textit{discriminator}, $f_{\phi}: \mathcal{Z} \rightarrow [0,1]$, which takes the latent representation of either a normal or a generated abnormal sample as input, and outputs the probability of the input being an anomaly. 
\end{itemize}
During training, the discriminator, $f_\phi$, receives both perturbed (abnormal) representation, $\tilde{z}$, and original (normal) representation, $z$, and is jointly trained with the encoder, $f_{\theta}$, to distinguish between them, while the perturbators, $g_{\psi_{1}},\dots,g_{\psi_L}$, are adversarially trained to fool the discriminator.
By doing so, we want the discriminator to be capable of responding to actual anomalies presenting heterogeneous and subtly defective traits. 
In the following subsections, we will describe each component of our proposed method in detail.

\subsection{Generating Hard-to-distinguish Synthetic Anomalies}
One desirable attribute of the perturbed representations designed for anomaly generation lies in their capability to render the task of distinguishing normal samples from synthetic anomalies challenging. This, in turn, will encourage the encoder and the discriminator to acquire useful and distinctive information that specifically represents normal samples. 
To achieve this goal, our initial step is to guarantee that the magnitude of each perturbation remains small by minimizing its Euclidean norm, i.e., $\| \epsilon_{\ell}^{i} \|_2$. This prevents the occurrence of a trivial scenario where the discriminator can effortlessly differentiate generated abnormal samples from normal ones as the perturbator can merely increase the perturbation level. 

Furthermore, we generate two types of perturbations, each playing opposite roles: one indicating the proximity of normal representations that should be classified as normal, and the other indicating proximity, yet with sufficient distance from normal representations, that should be classified as abnormal. 
Let $\epsilon^{i}_{\ell} \sim g_{\psi_\ell}(x^{i})$ be an input-dependent perturbation generated by the $\ell$-th perturbator taking $x^{i}$ as an input. Then, we define $(\epsilon^{\pi(1)}_{\ell}, \epsilon^{\pi(2)}_{\ell}, \dots, \epsilon^{\pi(m)}_{\ell})$ an ordered set of perturbations given $m$ samples, $\{x^{i}\}_{i=1}^{m}$, where $\pi$ is a permutation function that arranges them in increasing order of magnitude (i.e., Euclidean norm) such that $\|\epsilon^{\pi(1)}_{\ell}\|_{2} \leq \|\epsilon^{\pi(2)}_{\ell} \|_{2} \leq \cdots \leq \|\epsilon^{\pi(m)}_{\ell}\|_{2}$. 
From the ordered set of $m$ perturbations, we set the first $K$ as augmentations applied to normal samples, denoted as $\mathcal{E}^{+}_\ell \defeq \{ \epsilon_{\ell}^{\pi(j)} \}_{j=1}^{K}$ and set the remaining $(m-K)$ as anomalous versions of perturbations, denoted as $\mathcal{E}^{-}_\ell \defeq \{ \epsilon_{\ell}^{\pi(j)} \}_{j=K+1}^{m}$. 
Then, we assign a pseudo label for the perturbed latent representation of $x^i$, i.e., $\tilde{z}^{i} = z^{i} + \epsilon^{i}_{\ell}$, as the following:
\begin{equation} \label{eq:pseudo_label}
    \tilde{y}^i_\ell =\begin{cases} 
        0~~\text{(normal)} &\text{for}~~\epsilon_{\ell}^{i} \in \mathcal{E}^{+}_\ell \\
        1~~\text{(abnormal)} &\text{for}~~\epsilon_{\ell}^{i} \in \mathcal{E}^{-}_\ell.
    \end{cases}
\end{equation}
Based on pseudo labels obtained through \eqref{eq:pseudo_label}, we train $f_{\theta}$, $f_{\phi}$, and $g_{\psi_{\ell}}$ in order to differentiate between synthetically generated normal and abnormal samples.

Utilizing both synthetically generated normal and abnormal samples has the following two advantages: i) Treating augmented latent representations as normal samples preserves the original normality even after applying perturbation, thereby enhancing generalization to unseen normal samples. ii) By aligning perturbations used for augmentation with those employed for anomaly generation, we make the synthetically generated anomalies hard to distinguish. Consequently, this renders the differentiation between the two perturbed representations more challenging, preventing the occurrence of trivial cases where anomalous perturbations are merely small yet recognizable patterns.

\subsection{Diversifying Synthetic Anomalies}

Another desirable attribute of the perturbed representations designed for anomaly generation is their ability to attain diverse representations. This is crucial in maintaining the robustness of both the encoder and the discriminator against unseen real abnormal samples during testing.
To achieve this goal, we introduce multiple perturbators denoted as $g_{\psi_1}, g_{\psi_2},\dots, g_{\psi_L}$ where $L$ indicates the number of perturbators. 
Each of these perturbators is directed to generate perturbations in directions distinct from those of the other perturbators, thereby diversifying latent representations for the synthetically generated anomalies. 
More specifically, let $\epsilon_{1}^{i}, \epsilon_{2}^{i},\dots, \epsilon_{L}^{i}$ represent the perturbations generated by the $L$ perturbators when provided with $x^{i}$ as input. We foster diversity among the perturbations produced by different perturbators by minimizing their cosine similarity, expressed as $\texttt{sim}(\epsilon_{\ell}^{i}, \epsilon_{k}^{i}) = \frac{\epsilon_{\ell}^{i}\top \epsilon_{k}^{i}}{\| \epsilon_{\ell}^{i} \|_{2} \| \epsilon_{k}^{i} \|_2}$ for $\ell \neq k$.

\subsection{Optimizing the Three Components}
We jointly train the three components of our unsupervised anomaly detection framework -- $f_\theta$, $f_\phi$, and $\{g_{\psi_\ell} \}_{\ell=1}^{L}$ (parameterized by $\theta$, $\phi$, and $\{ \psi_{\ell} \}_{\ell=1}^{L}$, respectively) -- based on the following objective:
\begin{equation}
    \label{eq:overall_objective}
    \underset{\theta, \phi, \{ \psi_\ell \}_{\ell=1}^L }{\text{minimize}} \quad  \frac{1}{N_{tr}} \sum_{i=1}^{N_{tr}} ~ \mathcal{L}_{ce}^{i} + \lambda_1 \mathcal{L}_{norm}^{i} + \lambda_2 \mathcal{L}_{div}^{i} 
\end{equation}
where $\lambda_1, \lambda_2 \geq 0$ are hyper-parameters chosen to balance among different loss functions. 

The aim of our objective in \eqref{eq:overall_objective} is to distinguish both original and augmented (with perturbations) normal samples from synthetically generated abnormal samples based on 
\begin{equation} \label{eq:loss_ce} \nonumber
    \begin{split}
        \mathcal{L}_{ce}^{i} \!= \texttt{CE}(f_\phi (f_\theta(x^i)), 0) +  \frac{1}{L} \sum_{\ell=1}^L \texttt{CE}(f_\phi(f_\theta (x^i) \!+\! \epsilon_{\ell}^i), \tilde{y}_\ell^i) 
    \end{split}
\end{equation}
where $\texttt{CE}(\hat{y}, y)$ indicates the cross-entropy between $\hat{y}$ and $y$. Concurrently, the objective in \eqref{eq:overall_objective} encourages the synthetically generated anomalies to be diversified and hard to distinguish based on the following two loss functions:
\begin{equation} \label{eq:loss_others} \nonumber
    \mathcal{L}_{norm}^{i} \!= \frac{1}{L} \sum_{\ell=1}^L  \| \epsilon_{\ell}^i \|_2,~~    \mathcal{L}_{div}^{i}  \!= \frac{1}{L(L-1)} \!\sum_{\ell \neq k} \texttt{sim}(\epsilon_{\ell}^i, \epsilon_{k}^i).
\end{equation}
Here, $\mathcal{L}_{norm}^{i}$ ensures that the perturbations maintain a small magnitude, while $\mathcal{L}_{div}^{i}$ introduce diversity among the perturbations generated by different perturbators. The pseudo-code for the overall learning process is described in Algorithm \ref{alg:DRAG_train}.

Ultimately, we employ the trained encoder and discriminator pair as a score function, $S = f_{\phi} \circ f_{\theta}$, to detect whether a new data sample $x^{*}$ is classified as normal (i.e., $f_{\phi}(f_{\theta}(x^{*})) \leq \delta $) or abnormal (i.e., $f_{\phi}(f_{\theta}(x^{*})) > \delta$), based on a predefined threshold $\delta$ (e.g., $0.5$).

\subsection{Extension to Semi-Supervised Learning}
Suppose we are provided with a set of $N_{tr}^{s}$ (known) real anomaly samples, denoted as $\mathcal{D}_{tr}^{s}=\{(x^i, 1)\}_{i=1}^{N_{tr}^{s}}$. To harness such valuable information about the abnormal distribution, we make a straightforward extension of our objective in \eqref{eq:overall_objective} by augmenting it with a cross-entropy loss for the known anomaly samples as $\frac{1}{N_{tr}^{s}} \sum_{j=1}^{N_{tr}^{s}} \mathcal{L}_{aug}^{j}$, 
where $\mathcal{L}_{aug}^{j} = \texttt{CE}(f_{\phi}(f_{\theta}(x^{j})), 1)$.
% \begin{equation} \nonumber
%     \mathcal{L}_{aug}^{j} = \texttt{CE}(f_{\phi}(f_{\theta}(x^{j})), 1).
% \end{equation}
We will further present experimental results in the Experiments section, demonstrating that our unsupervised anomaly detection framework shows significant improvement through the augmentation of the loss function, which incorporates information about the known real anomaly samples.

%%%%%%%%%%%%%%%%%%%%%%%%%%%%%%%%%%%%%
% Main image results total (CIFAR-10 & FMNIST)
\begin{table*}[t]
\caption{The AUC performance (\%) on `one-vs.-rest' experiments for the two image datasets (i.e., CIFAR-10 and FMNIST). 
We report the mean and standard deviation of our method based on five randomly repeated tests. Here, ${*}$ indicates that the results are achieved by running experiments based on the official source code. The best two methods are highlighted in \textbf{bold}. For the benchmarks, we include the standard deviation reported in their published papers. If the standard deviation is not provided, we refrain from conducting additional random runs to maintain consistency with their optimal performance.}
\centering
\resizebox{\textwidth}{!}{%
\begin{tabular}{lcccccccccc}
% \hline
\textbf{CIFAR-10} & \multicolumn{1}{l}{} & \multicolumn{1}{l}{} & \multicolumn{1}{l}{} & \multicolumn{1}{l}{} & \multicolumn{1}{l}{} & \multicolumn{1}{l}{} & \multicolumn{1}{l}{} & \multicolumn{1}{l}{} & \multicolumn{1}{l}{} & \multicolumn{1}{l}{} \\ \hline
\multicolumn{1}{l|}{\begin{tabular}[c]{@{}l@{}}\textbf{Normal       Class}\end{tabular}} & \textbf{Airplane} & \textbf{Automobile} & \textbf{Bird} & \textbf{Cat} & \textbf{Deer} & \textbf{Dog} & \textbf{Frog} & \textbf{Horse} & \textbf{Ship} & \textbf{Truck} \\ \hline
\multicolumn{1}{l|}{\textbf{OCSVM}} & 61.6 & 63.8 & 50.0 & 55.9 & 66.0 & 62.4 & 74.7 & 62.6 & 74.9 & 75.9 \\
\multicolumn{1}{l|}{\textbf{iForest}} & 66.1 & 43.7 & 64.3 & 50.5 & \textbf{74.3} & 52.3 & 70.7 & 53.0 & 69.1 & 53.2 \\
\multicolumn{1}{l|}{\textbf{DAE}} & 41.1 & 47.8 & 61.6 & 56.2 & 72.8 & 51.3 & 68.8 & 49.7 & 48.7 & 37.8 \\
\multicolumn{1}{l|}{\textbf{DSVDD}} & 61.7 (4.1) & 65.9 (2.1) & 50.8 (0.8) & 59.1 (1.4) & 60.9 (1.1) & 65.7 (2.5) & 67.7 (2.6) & 67.3 (0.9) & 75.9 (1.2) & 73.1 (1.2) \\
\multicolumn{1}{l|}{\textbf{DAGMM}} & 41.4 & 57.1 & 53.8 & 51.2 & 52.2 & 49.3 & 64.9 & 55.3 & 51.9 & 54.2 \\
\multicolumn{1}{l|}{\textbf{ADGAN}} & 63.2 & 52.9 & 58.0 & 60.6 & 60.7 & 65.9 & 61.1 & 63.0 & 74.4 & 64.2 \\
\multicolumn{1}{l|}{\textbf{OCGAN}} & 75.7 & 53.1 & 64.0 & 62.0 & 72.3 & 62.0 & 72.3 & 57.5 & 82.0 & 55.4 \\
\multicolumn{1}{l|}{\textbf{TQM}} & 40.7 & 53.1 & 41.7 & 58.2 & 39.2 & 62.6 & 55.1 & 63.1 & 48.6 & 58.7 \\
\multicolumn{1}{l|}{\textbf{HRN}} & 77.3 (0.2) & 69.9 (1.3) & 60.6 (0.3) & 64.4 (1.3) & 71.5 (1.0) & 67.4 (0.5) & 77.4 (0.2) & 64.9 (1.1) & \textbf{82.5 (0.2)} & 77.3 (0.9) \\
\multicolumn{1}{l|}{\textbf{DROCC}} & 81.7 (0.2) & 76.7 (1.0) & 66.7 (1.0) & \textbf{67.1 (1.5)} & 73.6 (2.0) & \textbf{74.4 (2.0)} & 74.4 (1.0) & 71.4 (0.2) & 80.0 (1.7) & 76.2 (0.7) \\
\multicolumn{1}{l|}{\textbf{PLAD}} & \textbf{82.5 (0.4)} & \textbf{80.8 (0.9)} & \textbf{68.8 (1.2)} & 65.2 (1.2) & 71.6 (1.1) & 71.2 (1.6) & \textbf{76.4 (1.9)} & \textbf{73.5 (1.0)} & 80.6 (1.8) & \textbf{80.5 (1.3)} \\ \hline
\multicolumn{1}{l|}{\textbf{\proposed-variant}} & 75.0 (5.8) & 76.4 (3.6) & 64.3 (2.4) & 63.1 (3.9) & 70.4 (1.5) & 63.4 (2.8) & 70.1 (2.5) & 70.0 (3.3) & 75.4 (2.3) & 79.3 (0.9) \\
\multicolumn{1}{l|}{\textbf{\proposed}} & \textbf{83.6 (0.4)} & \textbf{81.1 (0.6)} & \textbf{71.2 (1.4)} & \textbf{66.2 (1.0)} & \textbf{76.3 (1.0)} & \textbf{72.2 (1.8)} & \textbf{82.2 (1.6)} & \textbf{79.3 (0.9)} & \textbf{85.8 (0.8)} & \textbf{84.2 (0.6)} \\ \hline
 & \multicolumn{1}{l}{} & \multicolumn{1}{l}{} & \multicolumn{1}{l}{} & \multicolumn{1}{l}{} & \multicolumn{1}{l}{} & \multicolumn{1}{l}{} & \multicolumn{1}{l}{} & \multicolumn{1}{l}{} & \multicolumn{1}{l}{} & \multicolumn{1}{l}{} \\ 
\textbf{FMNIST} & \multicolumn{1}{l}{} & \multicolumn{1}{l}{} & \multicolumn{1}{l}{} & \multicolumn{1}{l}{} & \multicolumn{1}{l}{} & \multicolumn{1}{l}{} & \multicolumn{1}{l}{} & \multicolumn{1}{l}{} & \multicolumn{1}{l}{} & \multicolumn{1}{l}{} \\ \hline
\multicolumn{1}{l|}{\begin{tabular}[c]{@{}l@{}}\textbf{Normal   Class}\end{tabular}} & \textbf{T-shirt} & \textbf{Trouser} & \textbf{Pullover} & \textbf{Dress} & \textbf{Coat} & \textbf{Sandal} & \textbf{Shirt} & \textbf{Sneaker} & \textbf{Bag} & \textbf{\begin{tabular}[c]{@{}c@{}}Ankle       boot\end{tabular}} \\ \hline
\multicolumn{1}{l|}{\textbf{OCSVM}} & 86.1 & 93.9 & 85.6 & 85.9 & 84.6 & 81.3 & 78.6 & 97.6 & 79.5 & 97.8 \\
\multicolumn{1}{l|}{\textbf{iForest}} & 91.0 & 97.8 & 87.2 & 93.2 & 90.5 & 93.0 & 80.2 & 98.2 & 88.7 & 95.4 \\
\multicolumn{1}{l|}{\textbf{DAE}} & 86.7 & 97.8 & 80.8 & 91.4 & 86.5 & 82.1 & 73.8 & 97.7 & 78.2 & 96.3 \\
\multicolumn{1}{l|}{\textbf{DSVDD$^{*}$}} & 85.2 (0.6) & 97.8 (0.2) & 84.0 (1.0) & 88.8 (0.9) & 85.0 (1.0) & 89.5 (0.3) & 78.2 (0.6) & 96.8 (0.2) & 92.2 (2.0) & 97.1 (0.3) \\
\multicolumn{1}{l|}{\textbf{DAGMM}} & 42.1 & 55.1 & 50.4 & 57.0 & 26.9 & 70.5 & 48.3 & 83.5 & 49.9 & 34.0 \\
\multicolumn{1}{l|}{\textbf{ADGAN}} & 89.9 & 81.9 & 87.6 & 91.2 & 86.5 & 89.6 & 74.3 & 97.2 & 89.0 & 97.1 \\
\multicolumn{1}{l|}{\textbf{OCGAN}} & 85.5 & 93.4 & 85.0 & 88.1 & 85.8 & 88.5 & 77.5 & 93.9 & 82.7 & 97.8 \\
\multicolumn{1}{l|}{\textbf{TQM}} & 92.2 & 95.8 & 89.9 & 93.0 & 92.2 & 89.4 & \textbf{84.4} & 98.0 & \textbf{94.5} & 98.3 \\
\multicolumn{1}{l|}{\textbf{HRN}} & 92.7 (0.0) & 98.5 (0.1) & 88.5 (0.1) & 93.1 (0.1) & 92.1 (0.1) & 91.3 (0.4) & 79.8 (0.1) & \textbf{99.0 (0.0)} & \textbf{94.6 (0.1)} & 98.8 (0.0) \\
\multicolumn{1}{l|}{\textbf{DROCC$^{*}$}} & 89.0 (1.8) & 97.6 (0.9) & 85.3 (2.7) & 91.6 (1.9) & 88.1 (2.9) & 95.4 (1.3) & 79.8 (0.5) & 95.0 (0.9) & 83.7 (3.5) & 97.8 (0.7) \\
\multicolumn{1}{l|}{\textbf{PLAD}} & \textbf{93.1 (0.5)} & \textbf{98.6 (0.2)} & \textbf{90.2 (0.7)} & \textbf{93.7 (0.6)} & \textbf{92.8 (0.8)} & \textbf{96.0 (0.4)} & 82.0 (0.6) & 98.6 (0.3) & 90.9 (1.0) & \textbf{99.1 (0.1)} \\ \hline
\multicolumn{1}{l|}{\textbf{\proposed-variant}} & 92.6 (1.0) & 97.7 (1.2) & 87.2 (1.7) & 90.9 (1.4) & 91.5 (0.4) & 95.5 (0.6) & 81.5 (0.9) & 96.8 (0.8) & 88.4 (3.0) & 98.8 (0.3) \\
\multicolumn{1}{l|}{\textbf{\proposed}} & \textbf{95.5 (0.2)} & \textbf{99.3 (0.1)} & \textbf{92.0 (1.0)} & \textbf{94.9 (1.2)} & \textbf{94.1 (0.5)} & \textbf{95.5 (0.4)} & \textbf{83.8 (1.0)} & \textbf{99.0 (0.0)} & \textbf{94.5 (1.4)} & \textbf{99.5 (0.1)} \\ \hline
\end{tabular}%
}
\label{Table:image}
\end{table*}
%%%%%%%%%%%%%%%%%%%%%%%%%%%%%%%%%%%%%

\section{Experiments}
In this section, we evaluate our proposed method, \proposed, across various real-world datasets commonly used for anomaly detection tasks, encompassing two distinct modalities: image and non-image tabular data. As elaborated in Section 4, the anomaly score is computed during inference by passing the input data sample through the trained encoder, followed by the trained discriminator. 

\paragraph{Implementation.}
The parameters for the encoder, discriminator, and the set of perturbators -- i.e., $(\theta, \phi, \{\psi_{\ell}\}_{\ell=1}^{L})$ -- are jointly optimized via Adam optimizer \cite{kingma2014adam} with learning rates of $5\times 10^{-3}$ for $(\theta, \phi)$ and $1\times 10^{-5}$ for $\{\psi_{\ell}\}_{\ell=1}^{L}$, respectively. This remains consistent across all types of datasets. We use grid search to obtain the best hyper-parameters that varied across the datasets. The balancing coefficients $\lambda_1, \lambda_2$ are chosen from $\{10^{-4}, 10^{-3}, 10^{-2}, 10^{-1}, 10^{0} \}$. The number of augmented versions of samples $K$ and the number of perturbators $L$ are chosen from $\{ 30, 50, 100 \}$ and $\{ 3, 5, 10 \}$, respectively. We fix the subset size $m$ as $512$.

We implement the three components of \proposed~using different network architectures for image and tabular datasets, respectively: For image datasets, we employ a LeNet-based CNN for the encoder, which is also used in \cite{PLAD,DROCC}, an MLP for the discriminator, and a 2D-CNN for the perturbators.  For tabular datasets, we employ MLPs for both the encoder and the discriminator, and a 1D-CNN for the perturbators. We choose CNN for the perturbators motivated by recent success in generating domain-agnostic transformations using trainable perturbations \cite{VIEWMAKER}. More specifically, for the image dataset, each perturbator is a stochastic function that randomly generates perturbations by integrating a random noise channel into the input feature -- i.e., from $\mathbb{R}^{W\times H\times C}$ to $\mathbb{R}^{W\times H\times (C+1)}$ -- during forward computation. Here, $W$, $H$, and $C$ denote the width, height, and number of channels, respectively. For the tabular dataset, each perturbator randomly generates stochastic perturbations by incorporating a random noise channel into the input features (i.e., $\mathbb{R}^{d} \rightarrow \mathbb{R}^{d+1}$), similar to the settings used for the image datasets.

\paragraph{Benchmarks.}
Throughout the experiments, we compare the anomaly detection performance of \proposed~against 16 state-of-the-art anomaly detection methods: 3 shallow approaches such as \textbf{OCSVM} \cite{scholkopf1999support}, \textbf{iForest} \cite{liu2012isolation}, \textbf{LOF} \cite{breunig2000lof}, 13 deep learning-based approaches such as \textbf{DAE} \cite{DAE}, \textbf{DSVDD} \cite{DSVDD}, \textbf{DAGMM} \cite{DAGMM}, \textbf{ADGAN} \cite{ADGAN}, \textbf{OCGAN} \cite{OCGAN}, \textbf{TQM} \cite{TQM}, \textbf{HRN} \cite{HRN}, \textbf{DROCC} \cite{DROCC}, \textbf{PLAD} \cite{PLAD}, \textbf{GOAD} \cite{bergman2020goad}, \textbf{NeuTraLAD} \cite{qiu2021neural}, and \textbf{DeepiForest} \cite{xu2023deep}, and 1 semi-supervised approach, \textbf{DeepSAD} \cite{DeepSAD}. For all benchmarks, when experimental results under the same conditions (i.e., data preprocessing, evaluation protocol) are available in the existing publications, we explicitly state them; if not, we run their official source code according to the optimal hyper-parameter settings described in those papers.

% New tabular dataset table 
\begin{table}[t!]
\caption{The F1 score (\%) on the five tabular datasets (i.e., Thyroid, Pageblocks, Shuttle, KDD, and KDDRev). 
We report the mean and standard deviation of our method based on five randomly repeated tests. Here, ${*}$ indicates that the results are achieved by running experiments based on the official source code.  
The best two methods are highlighted in \textbf{bold}.}
\centering
\resizebox{\columnwidth}{!}{%
\begin{tabular}{l|ccccc}
\hline
\textbf{Method}      & \textbf{Thyroid}   & \textbf{KDD}       & \textbf{KDDRev}    & \textbf{Pageblocks$^{*}$} & \textbf{Shuttle$^{*}$}   \\ \hline
\textbf{OCSVM}       & 38.9               & 79.5               & 83.2               & 56.4(0.0)           & 61.5(0.0)          \\
\textbf{iForest}     & 52.7               & 90.7               & 90.6               & 45.1(9.0)           & 28.0(0.8)          \\
\textbf{LOF}         & 52.7(0.0)          & 83.8(5.2)          & 81.6(3.6)          & 69.4(0.0)           & 53.8(0.0)          \\
\textbf{DAGMM}       & 47.8               & 93.7               & 93.8               & 65.6(2.9)           & 67.7(28.1)         \\
\textbf{GOAD}        & 74.5(1.1)          & \textbf{98.4(0.2)} & 98.9(0.3)          & 67.5(1.4)           & 58.5(3.4)          \\
\textbf{NeuTraLAD}   & \textbf{76.8(1.9)} & \textbf{99.3(0.1)} & \textbf{99.1(0.1)} & 61.1(2.5)           & \textbf{81.5(7.8)} \\
\textbf{DROCC$^{*}$}       & 74.2(1.5)          & 65.4 (10.9)          & 95.0(1.7)          & \textbf{73.9(3.5)}  & 60.0(5.8)          \\
\textbf{PLAD$^{*}$}        & 76.6(0.6)          & 66.1 (2.5)          & 94.5(1.3)          & 58.5(7.1)           & 70.8(7.5)          \\
\textbf{DeepiForest$^{*}$} & 67.3(1.7)          & 74.1(5.0)          & 50.0(0.0)          & 48.8(0.9)           & 40.1(2.6)          \\ \hline
\textbf{\proposed}        & \textbf{86.9(0.9)} & 98.2(0.4)          & \textbf{98.9(0.1)} & \textbf{78.8(1.5)}  & \textbf{83.1(5.8)} \\ \hline
\end{tabular}%
}
\label{tab:Tabular_results}
\end{table}
%%%%%%%%%%%%%%%%%%%%%%%%%%%%%%%%%%%%%

\subsection{Experiments on Image Data}

\paragraph{Dataset Description.}  
We utilize two image datasets commonly used in anomaly detection: FMNIST\footnote{\url{https://github.com/zalandoresearch/fashion-mnist}} \cite{FMNIST} and CIFAR-10\footnote{\url{https://www.cs.toronto.edu/~kriz/cifar.html}} \cite{CIFAR10}. Each dataset consists of ten distinct class labels, where one class is designated as normal samples and the other nine classes are treated as abnormal samples. 
More detailed descriptions are as follows:

\begin{itemize}[leftmargin=1.5em]
    \item FMNIST:~~This dataset is a collection of images with information about 10 classes such as t-shirts, sandals, bags, and others. Among the 70,000 images, we spare 60,000 images (i.e., 6,000 images for each class) for training and the remaining 10,000 samples for testing. Thus, for experiments on each normal class, we use 6,000 images for training anomaly detection methods. 
    
    \item CIFAR-10:~~This dataset comprises images with information about 10 classes such as airplane, cat, deer, and others.  
    Out of the 60,000 images, we split the dataset into 50,000 images for the training set and the remaining 10,000 images for the testing set. The CIFAR-10 dataset exhibits a broader distributional diversity with respect to objects and backgrounds when compared to the FMNIST dataset. Such complexity of visual features in the CIFAR-10 dataset can obscure the boundaries between normal samples and abnormal samples, making it more challenging for anomaly detection.
\end{itemize}

\paragraph{Results.} We conduct experiments on the image datasets following the standard protocol \cite{DSVDD} in one-class classification. 
In Table \ref{Table:image}, we compare the anomaly detection performance of \proposed~with the benchmarks with respect to the area under the ROC curve (AUC) on the image datasets. 
Our proposed method significantly outperforms or provides comparable results to the benchmarks on both datasets, including cutting-edge anomaly detection methods. 
In the CIFAR-10 dataset, \proposed~achieves the best performance for all the classes except for the two classes (i.e., `cat' and `dog'), where our method performs the second best. Remarkably, in four of the classes (i.e., `frog', `horse', `ship', and `truck'), \proposed~provides substantial performance gain over the best-performing benchmarks by more than 3\%. 
In the FMNIST dataset, our method outperforms (i.e., achieves the best performance in seven classes) or provides comparable performance (i.e., ranks second best in three classes with a marginal performance gap of below 1\%) over the benchmarks.
Through these experiments, we demonstrate that \proposed~effectively learns a decision boundary capable of distinguishing diverse image outliers, outperforming methods based on learning a single perturbator, i.e., PLAD and DROCC, in most cases.

\subsection{Experiments on Tabular Data}

\paragraph{Dataset Descriptions.} 
We use five real-world datasets: Thyroid \cite{thyroid}\footnote{\url{https://archive.ics.uci.edu/dataset/102/thyroid+disease}}, Pageblocks\footnote{\label{foot:pageblocks} \url{https://www.dbs.ifi.lmu.de/research/outlier-evaluation/DAMI/}} \cite{pageblocks}, Shuttle\textsuperscript{\ref{foot:pageblocks}} \cite{asuncion2007uci}, KDD\footnote{\label{foot:kdd} \url{http://kdd.ics.uci.edu/databases/kddcup99/kddcup99.html}}, and KDDRev\textsuperscript{\ref{foot:kdd}} \cite{kdd}. Thyroid is a small-scale dataset with fewer features (i.e., $6$) and samples (i.e., $3772$). 
Pageblocks and Shuttle are datasets sourced from the anomaly benchmark study \cite{campos2016evaluation}, both of which have small sample sizes (i.e., $5473$ and $1013$, respectively) comprised solely of numerical attributes (i.e., $10$ and $9$, respectively).
In contrast, the KDD and KDDRev are more complex as these datasets comprise both categorical and continuous variables (i.e., $7$ and $34$, respectively) with a relatively large number of samples (i.e., $494021$ and $121597$, respectively). 
More detailed descriptions are as follows:

\begin{itemize}[leftmargin=1.5em]
    \item Thyroid:~~This dataset comprises patients with thyroid disease whose outcomes are divided into three distinct classes depending on the presence of hypothyroidism. For the task of anomaly detection, we set samples with the least frequent (i.e., the minority) class, which is `hyperfunction', as abnormal samples.

    \item KDD:~~This dataset is a subset of the KDDCUP99, which comprises 10\% of the overall samples from KDDCUP99. The dataset consists of the normal and 4 simulated attack types (i.e., denial of service, unauthorized access from a remote machine, unauthorized access from local superuser, and probing). In this dataset, the normal class consists of only 20\% of the entire dataset and therefore is treated as abnormal, while the four attack types are treated as normal.

    \item KDDRev:~~This dataset has the same data samples with those of the KDD dataset. However, to be more consistent with the actual use case, the original `normal' class is kept as normal samples, while a random subset is extracted from the intrusion classes to serve as abnormal samples. 

    \item Pageblocks:~~This dataset contains information about different types of blocks in document pages. Blocks containing text are labeled as inliers, while those with non-text content are designated as outliers. 

    \item Shuttle:~~Initially, this dataset is a multi-class classification dataset included in the UCI machine learning repository. In the anomaly benchmark study \cite{campos2016evaluation}, classes $2, 3, 5, 6,$ and $7$ are reclassified as outliers, while class label $1$ forms an inlier.
\end{itemize}

\paragraph{Results.}
For a fair comparison, we follow the evaluation protocol and data preprocessing (e.g., train-test split, normalization) as conducted in the prior works \cite{DAGMM,bergman2020goad,qiu2021neural}. 
In Table \ref{tab:Tabular_results}, we compare the anomaly detection performance of \proposed~with the benchmarks with respect to the F1 score. 
As shown in Table \ref{tab:Tabular_results}, \proposed~consistently matches or outperforms the state-of-art methods across all the evaluated datasets. Specifically, in the Thyroid dataset, DHAG achieves a remarkable performance improvement over the best-performing benchmark, NeuTraLAD, with a notable margin (i.e., over 10\%) in terms of the average F1 score. 
In the Pageblocks and Shuttle datasets, our method also notably outperforms the second-best methods, DROCC and NeuTraLAD.
This demonstrates that even in the tabular dataset with a relatively limited sample size, \proposed~can effectively learn a decision boundary that can better describe the underlying representation of normal data. In the KDD and KDDRev datasets, \proposed~consistently provides a prediction performance comparable to the top-performing benchmarks, GOAD and NeuTraLAD, that are tailored for (non-image) tabular data.

%%%%%%%%%%%%%%%%%%%%%%%%%%%%%%%%%%%%%
% TSNE visualization
\begin{figure*}[t]
\centering

\subfloat[$K=0, L=10$]{%
  \includegraphics[width=5.6cm, trim=2em 8em 0em 15em, clip]{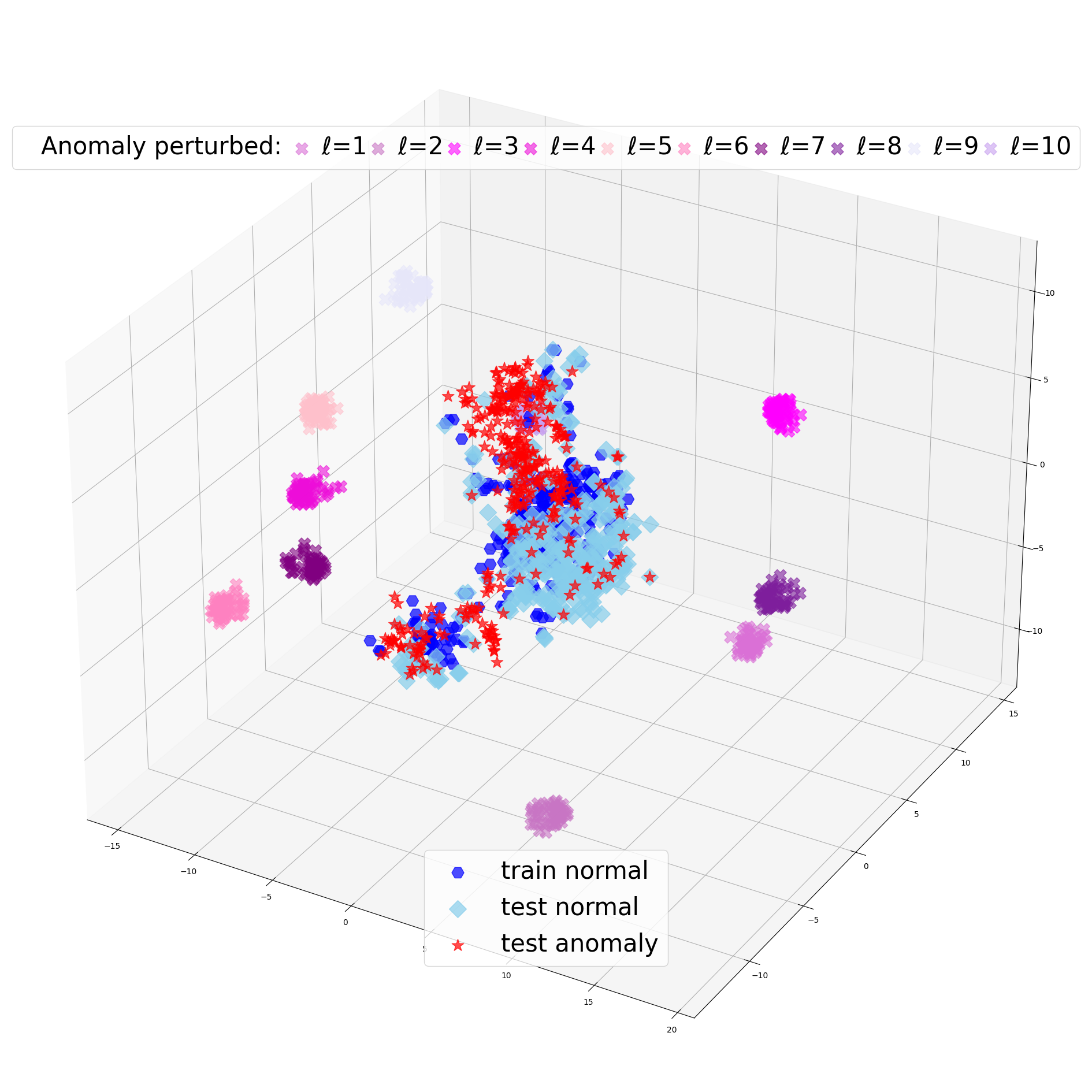}%
  \label{fig:latent_visualization_by_K0}%
}
\subfloat[$K=50, L=10$]{%
  \includegraphics[width=5.6cm, trim=2em 8em 0em 15em, clip]{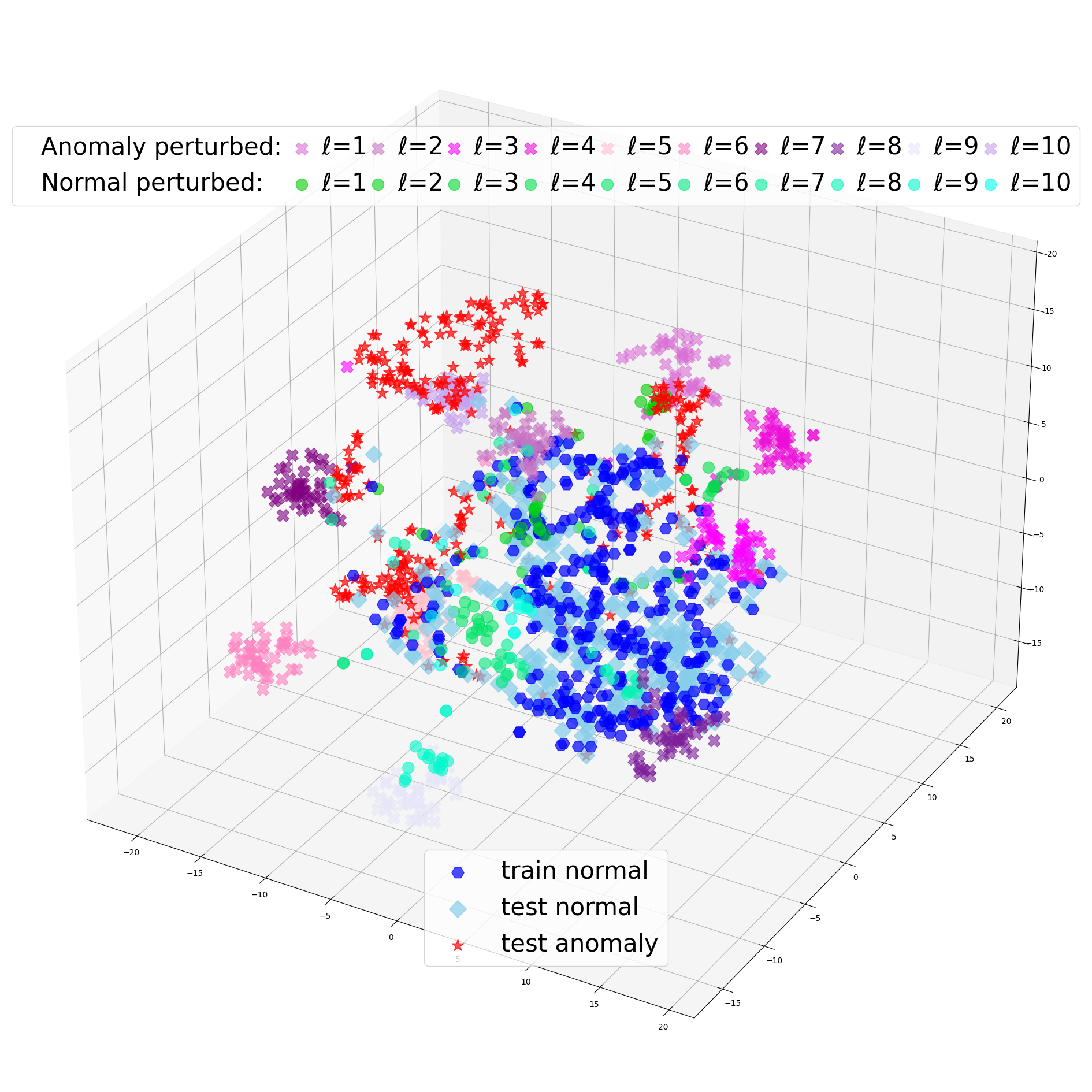}%
  \label{fig:latent_visualization_by_K50}%
}
\subfloat[$K=50, L=1$]{%
  \includegraphics[width=5.6cm, trim=2em 8em 0em 15em, clip]{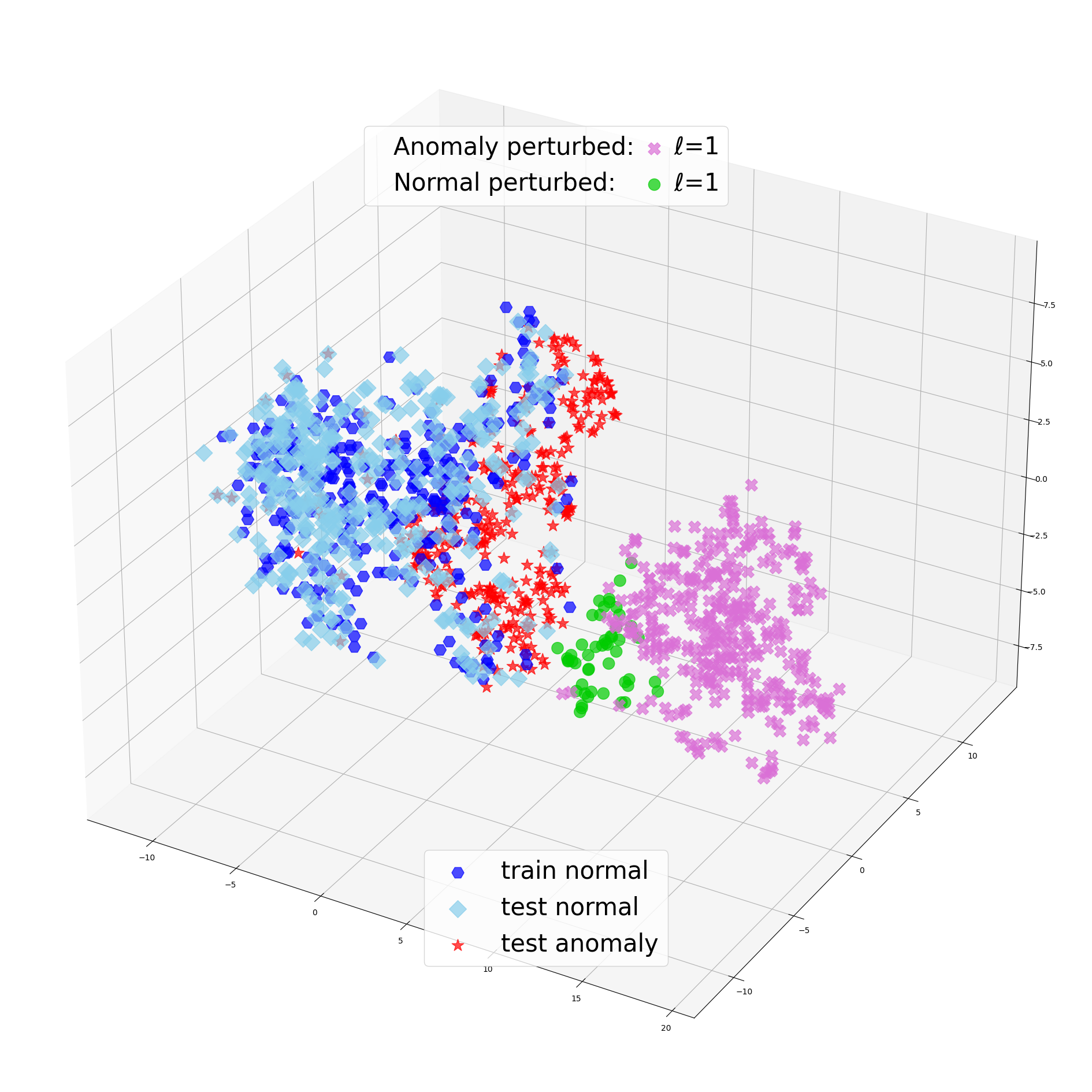}%
  \label{fig:latent_visualization_by_Nump1}%
}

\caption{The t-SNE visualization for `T-shirt' class from the FMNIST dataset. The visualizations contain five different sample types: perturbed counterparts of normal samples with $\tilde{y}_{\ell} = 1$ (\textit{Anomaly perturbed}) and those with $\tilde{y}_{\ell} = 0$ (\textit{Normal perturbed}), test anomaly samples (\textit{test anomaly}), test normal samples (\textit{test normal}), and train normal samples (\textit{train normal}).}

\label{fig:tsne_visualization}
\end{figure*}
%%%%%%%%%%%%%%%%%%%%%%%%%%%%%%%%%%%%%

\subsection{Ablation Study} 
In this subsection, we further provide the effect of two key factors of \proposed~-- i) the number of augmented versions of normal samples for each perturbator (i.e., $K$), and ii) the number of perturbators (i.e., $L$) -- using the image datasets. 

In Figure \ref{fig:tsne_visualization}, we present the t-SNE visualizations of the learned representations for the FMNIST dataset with three different scenarios: $(K=0, L=10)$, $(K=50, L=10)$, and $(K=50, L=1)$. 
Comparisons across the three scenarios offer valuable insights into the benefits of leveraging the two key factors. First, as shown in Figures \ref{fig:tsne_visualization}(a) and \ref{fig:tsne_visualization}(b), each perturbator generates perturbations that span directions different from those of other perturbators. However, the perturbed representations deviate significantly from the normal representations when the augmented normal samples are not generated simultaneously. This, in turn, makes it trivial for the discriminator to differentiate between normal samples and synthetically generated anomalies, resulting in a performance drop in detecting unseen anomalies during testing.  
Second, Figures \ref{fig:tsne_visualization}(b) and \ref{fig:tsne_visualization}(c) show that although there is limited overlap between synthetically generated anomalies and the real (unseen) anomalies, training discriminator using diverse latent representations of synthetic anomalies helps to learn decision boundaries that more effectively capture the distribution of normal samples. This enhancement contributes to improved generalization for unseen anomalies.

\begin{figure}[t]
\captionsetup[subfigure]{justification=centering}
\centering
\subfloat[AUC vs $K$]{%
  \includegraphics[width=4.1cm, trim=0 0 0 1em, clip]{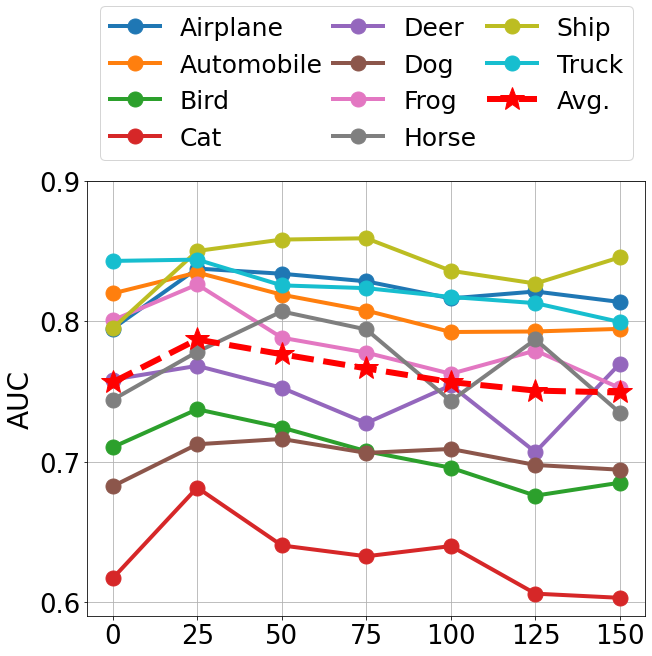}%
  \label{numberK}%
}\
\subfloat[AUC vs $L$]{%
  \includegraphics[width=4.1cm, trim=0 0 0 1em, clip]{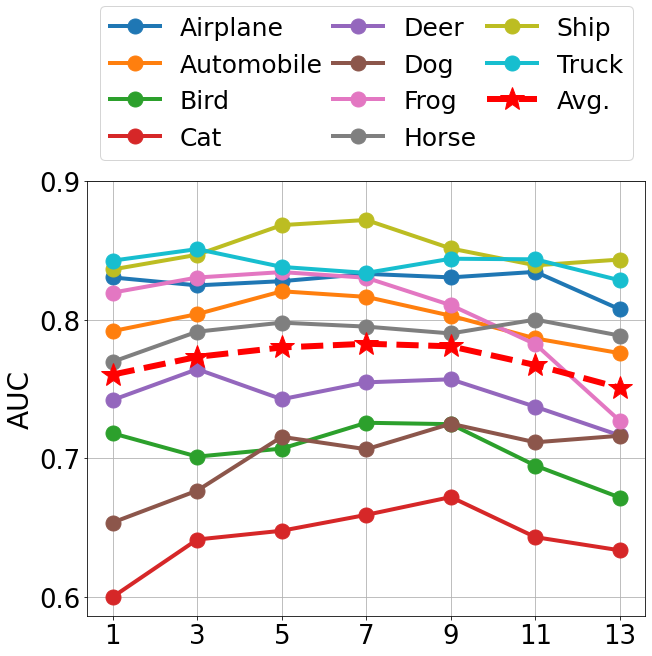}%
  \label{numberP}%
}
\caption{The average AUC for the CIFAR-10 dataset (both for each class and for overall average) with (a) varying $K$ with fixed $L$ and (b) varying $L$ with fixed $K$.}
\label{fig:ablation_study}

\end{figure}

In Figure \ref{fig:ablation_study}, we provide a sensitive analysis to see the influence of the two key factors with respect to the anomaly detection performance using the CIFAR-10 dataset. More specifically, we report the average AUC performance by varying $K$ while fixing $L$ in Figure \ref{fig:ablation_study}(a) and by varying $L$ while fixing $K$ in Figure \ref{fig:ablation_study}(b). In Figure \ref{fig:ablation_study}(a), we can observe that the performance is sub-optimal when the number of augmented normal samples is either too small (i.e., all synthetic anomalies) or too large (i.e., no synthetic anomalies). Also, Figure \ref{fig:ablation_study}(b) demonstrates that, in most cases, performance gradually improves as the number of perturbators increases up to a certain threshold, leveraging the effect of diversified synthetic anomalies. However, performance diminishes when the number of perturbators becomes excessively large, and the optimal value of $L$ can vary based on different normal classes.

%%%%%%%%%%%%%%%%%%%%%%%%%%%%%%%%%%%%%
% ablation study plot 
%%%%%%%%%%%%%%%%%%%%%%%%%%%%%%%%%%%%%

\subsection{Semi-Supervised Learning}  \label{subsection:semi_sup}
In this subsection, we further evaluate how well our proposed method can be extended to the semi-supervised setting, where a small number of true anomalies are available during training (See Section 4.4). In Table \ref{tab:semi-sup}, we compare the average AUC performance of \proposed~with DeepSVDD \cite{DSVDD} and its semi-supervised variant, DeepSAD \cite{DeepSAD}, on the two image datasets (i.e., FMNIST and CIFAR-10) and one tabular dataset (i.e., Thyroid). 
For the image datasets, we first set one out of the 10 classes as normal and set the remaining classes as abnormal. Then, we choose one class from the pool of 9 available anomaly classes to be the known anomaly. During training, we utilize samples from this known anomaly class, as described in Section 4.4. It is important to note that there are 9 distinct scenarios for each normal class, depending on which class is chosen as known anomalies. Overall, this experimental setting results in a total of 90 different scenarios based on the chosen normal and known anomaly classes. 
We change the ratio $\gamma=N_{tr}^s / (N_{tr}+N_{tr}^s)$ of known anomalies $(x^j, y^j=1)$, where $N_{tr}^{s}$ is the number of known anomalies. Hence, $\gamma=0$ implies the fully unsupervised setting. We assess the AUC performance by averaging all the possible 90 scenarios for each ratio of $\gamma$. 
% More specifically, we change the ratio $\gamma=N_{tr}^{s}/(N_{tr}+N_{tr}^{s})$ of known anomalies, where $N_{tr}^{s}$ is the number of known anomalies, in the overall data. Hence, $\gamma=0$ implies the fully unsupervised setting. Please see more details about the experiment setup in the Appendix. 
For the tabular dataset, we also conduct experiments by randomly selecting the ground-truth anomalous samples with a ratio of $\gamma$ and designating those samples as labeled samples, i.e., $(x^j, y^j=1)$. We report the mean and standard deviation based on five randomly repeated tests. Following the data preprocessing procedure described in the prior work \cite{DeepSAD}, we perform the random training/testing split of 60/40 while preserving the original proportion of anomalies in each set. Consequently, this leads to a discrepancy in the data settings from the experiments in Section 5.2 and Table \ref{tab:Tabular_results}, such as the proportion of anomalies in the test set decreasing from 4.8 \% to 2.5\%. This discrepancy leads to variations in some experimental outcomes, for instance, the performance of DHAG-unsup decreases from 86.9 to 81.1. 
Table \ref{tab:semi-sup} shows that \proposed~is capable of extracting information from known anomalous data samples during training, even when the number of the known anomalies is limited, resulting a significant performance improvement over the fully unsupervised setting.

\begin{table}[t]
\caption{The performance metric for image and tabular datasets in the semi-supervised setting with varying known-anomaly ratios $\gamma$. We report AUC (\%) for image datasets (CIFAR-10, FMNIST) and F1 score  (\%) for tabular dataset (Thyroid).}
\label{tab:semi-sup}
\resizebox{\columnwidth}{!}{%
\begin{tabular}{cccccc}
\hline
\textbf{Data} & \textbf{$\gamma$} & \textbf{\begin{tabular}[c]{@{}c@{}}\proposed-\\      unsup\end{tabular}} & \textbf{\begin{tabular}[c]{@{}c@{}}Deep\\      SVDD\end{tabular}} & \textbf{\begin{tabular}[c]{@{}c@{}}\proposed-\\      semisup\end{tabular}} & \textbf{\begin{tabular}[c]{@{}c@{}}Deep\\      SAD\end{tabular}} \\ \hline
\textbf{CIFAR-10} & 0.00 & 78.7   (6.2) & 60.9 (9.4) & 78.7 (6.2) & 60.9 (9.4) \\
\textbf{(Image-AUC)} & 0.01 &  &  & 80.4 (7.4) & 72.6 (7.4) \\
\textbf{} & 0.05 &  &  & 84.0 (6.6) & 77.9 (7.2) \\
\textbf{} & 0.10 &  &  & 85.2 (6.4) & 79.8 (7.1) \\
\textbf{} & 0.20 &  &  & 86.6 (6.1) & 81.9 (7.0) \\ \hline
\textbf{FMNIST} & 0.00 & 94.68   (4.34) & 89.2 (6.2) & 94.68 (4.34) & 89.2 (6.2) \\
\textbf{(Image-AUC)} & 0.01 &  &  & 95.61 (4.29) & 90.0   (6.4) \\
\textbf{} & 0.05 &  &  & 95.57 (4.27) & 90.5   (6.5) \\
\textbf{} & 0.10 &  &  & 95.52 (4.36) & 91.3   (6.0) \\
\textbf{} & 0.20 &  &  & 95.56 (4.27) & 91.0   (5.5) \\ \hline
\textbf{Thyroid} & 0.00 & 81.1 (2.9) & 49.5 (4.2) & 81.1 (2.9) & 49.5 (4.2) \\
\textbf{(Tabular-F1 score)} & 0.01 &  &  & 83.7 (2.9) & 77.9 (5.4) \\
\textbf{} & 0.02 &  &  & 85.8 (1.4) & 80.0 (4.3) \\ \hline
\end{tabular}%
}
\end{table}
%%%%%%%%%%%%%%%%%%%%%%%%%%%%%%%%%%%%%

%%%%%%%%%%%%%%%%%%%%%%%%%%%%%%%%%%%%%
% Feature space perturbation 
\begin{figure}[h]
    \centering
    \includegraphics[width=1.0\linewidth, trim=0em 0em 0em 0em, clip]{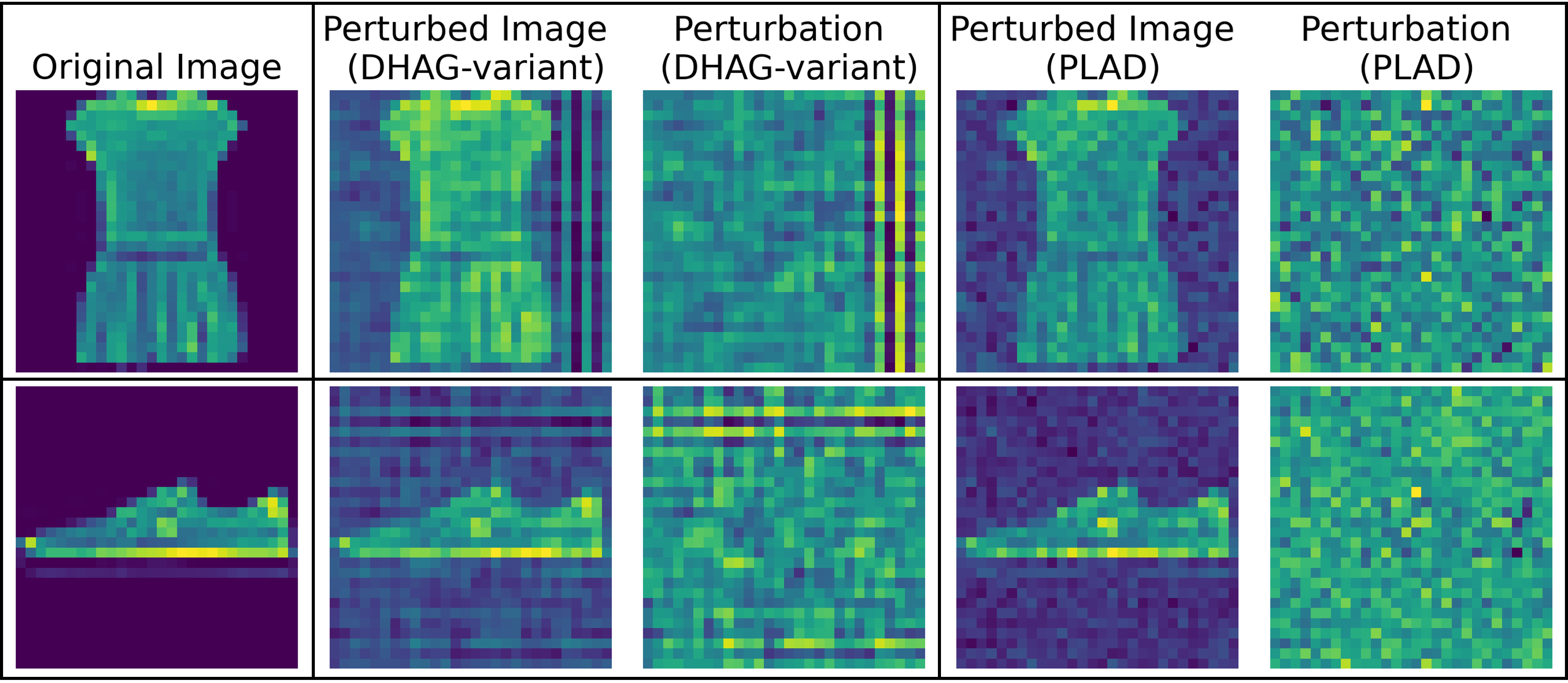}
    \caption{The visualization of images (i.e., FMNIST) before and after the application of feature-level perturbation generated by \proposed-variant and PLAD methods.}
    \label{fig:feature_space_perturbation}
\end{figure}
%%%%%%%%%%%%%%%%%%%%%%%%%%%%%%%%%%%%%

%%%%%%%%%%%%%%%%%%%%%%%%%%%%%%%%%%%%%
% Case study (DRAG vs iForest) 
\begin{figure}[h]
    \centering
    \includegraphics[width=1.0\linewidth, trim=0em 0em 0em 0em, clip]{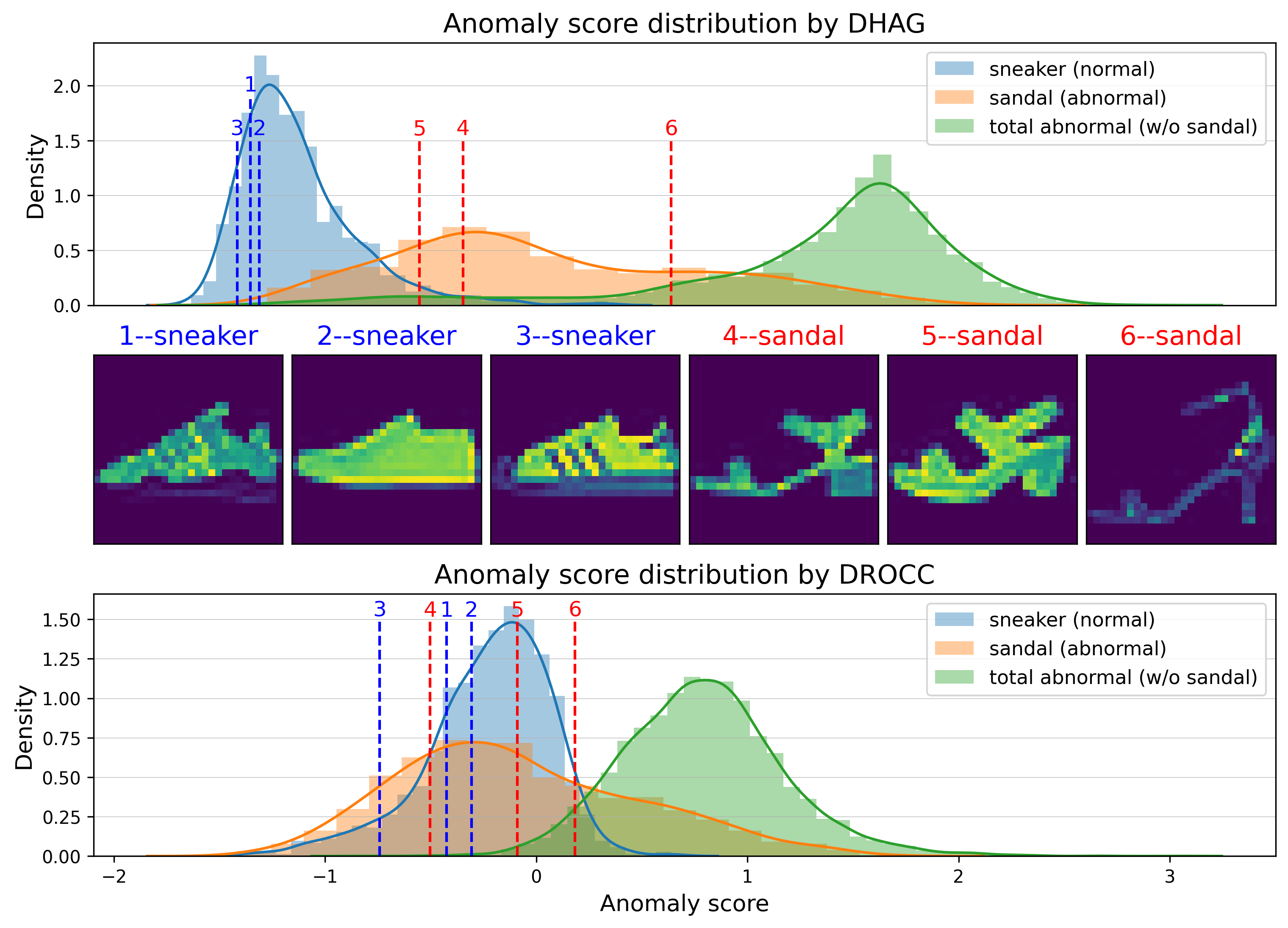}
    \caption{The visualization of test anomaly score distributions in the FMNIST dataset, where the `sneaker' class is designated as the normal training data, with a comparative analysis between \proposed~and DROCC. The figure also displays anomaly scores for some selected samples.}
    \label{fig:case_study}
\end{figure}
%%%%%%%%%%%%%%%%%%%%%%%%%%%%%%%%%%%%%

\subsection{Effects of the Latent Perturbations}
In this experiment, we investigate the effects of applying perturbations in the latent space by introducing a variant of \proposed. This variant incorporates perturbations directly into the input features, as opposed to applying perturbations into the latent representations. 
Figure \ref{fig:feature_space_perturbation} depicts some randomly selected images before and after applying the feature-level perturbations. The figure shows that the perturbed images contain easily distinguishable patterns in the background, making the auxiliary classification task (i.e., differentiating normal samples from their counterparts) much easier since neural networks (especially, the CNNs) tend to focus on easily distinguishable patterns such as texture rather than shape \cite{geirhos2018imagenettrained}. This, in turn, degrades the anomaly detection performance when compared to \proposed~as shown in Table \ref{Table:image}.

Now, we further investigate whether PLAD \cite{PLAD}, which is most closely related to our method, shows a similar tendency as perturbations are directly applied to the input features. Particularly, PLAD generates two types of outputs in an input-dependent way, which are directly multiplied and added to alter input features. 
In Figure \ref{fig:feature_space_perturbation}, we can observe that the perturbed images contain simple patterns, similar to salt and pepper noise, which are likely to be easily distinguished from normal samples. We believe that the presence of these easily distinguishable patterns could simplify the auxiliary task of distinguishing normal samples from generated anomalies. Consequently, this could lead to a decrease in anomaly detection performance.

\subsection{Case study}
We expand our comparative experiment by including an additional case study, specifically focusing on the scenario in which \proposed~successfully distinguishes between normal samples and the corresponding abnormal counterparts, whereas the perturbation-based baseline (i.e., DROCC) fails to do so, using the FMNIST dataset. 
We consider a one-class classification setting where we define the class `sneaker' as normal and all other classes as abnormal. Here, the anomaly detection methods are trained solely based on the normal samples with class `sneaker' during training. 

Figure \ref{fig:case_study} illustrates the distribution of test anomaly scores for samples with class `sneaker', `sandal', and all other classes (excluding `sandal'). This highlights the ability of our proposed method to discriminate between sneakers and sandals, particularly those that are \textit{hard to distinguish} due to their shared characteristics as footwear.
\proposed~provides anomaly scores with a clear distinction between normal and abnormal classes, particularly showing effective discrimination of the `sandal' class as abnormal.
In contrast, the decision boundary provided by DROCC is unclear because there exists a substantial overlap in the anomaly score distribution between the hard-to-distinguish `sandal' class and the normal `sneaker' class.
This result demonstrates that \proposed~is capable of distinguishing between normal samples exhibiting diverse characteristics and hard-to-distinguish abnormal samples, effectively reducing potential false alarms.

\section{Conclusion}
In this work, we introduce a novel unsupervised anomaly detection method, \proposed, that is trained based on an auxiliary classification task of distinguishing both original and augmented normal samples from synthetically generated abnormal samples. 
Our novel perturbation process encourages these synthetic anomalies to achieve two desirable attributes by simultaneously generating perturbations for both normal and abnormal transformations of given normal samples and by incorporating multiple perturbators spanning distinct perturbation directions. As a result, the auxiliary classification task becomes more challenging which improves distinguishing unseen abnormal samples from normal samples. 
Throughout the experiments, we evaluate \proposed~on both real-world image and tabular datasets, demonstrating the superiority of our method across different data types compared to the state-of-the-art benchmarks. The significance of our novel perturbation process is further validated by visualizing the learned representations using different variants of our perturbation process. 

\begin{acks}
We thank anonymous reviewers for many insightful comments and suggestions. This work was supported by the Institute of Information and communications Technology Planning \& Evaluation (IITP) funded by the Korea government (MSIT) under Grant 2021-0-01341 by the AI Graduate School Program (CAU) and 2019-0-00079 by the AI Graduate School Program (Korea University).
\end{acks}

%% The next two lines define the bibliography style to be used, and
%% the bibliography file.
% \bibliographystyle{ACM-Reference-Format}
% \balance
% \bibliography{DRAG}

%%% -*-BibTeX-*-
%%% Do NOT edit. File created by BibTeX with style
%%% ACM-Reference-Format-Journals [18-Jan-2012].

%%
%% If your work has an appendix, this is the place to put it.
% \appendix

\end{document}